\documentclass[11pt]{article}

\usepackage[letterpaper,margin=0.86in]{geometry}
\usepackage[T1]{fontenc}
\usepackage[utf8]{inputenc}
\IfFileExists{newtxtext.sty}{\usepackage{newtxtext,newtxmath}}{\usepackage{lmodern}}
\usepackage{microtype}
\usepackage{graphicx}
\usepackage{booktabs}
\usepackage{tabularx}
\usepackage{array}
\usepackage{multirow}
\usepackage{threeparttable}
\usepackage{amsmath}
\usepackage{siunitx}
\usepackage[font=small,labelfont=bf,labelsep=period]{caption}
\usepackage[numbers,sort&compress,super]{natbib}
\usepackage{lineno}
\usepackage{xcolor}
\usepackage{enumitem}
\usepackage{titlesec}
\usepackage{hyperref}

\definecolor{NatureTeal}{HTML}{15687F}
\definecolor{NatureAccent}{HTML}{E76F51}
\definecolor{Muted}{HTML}{667085}
\definecolor{LightRule}{HTML}{D0D7DE}

\titleformat{\section}{\large\bfseries}{\thesection}{0.65em}{}
\titleformat{\subsection}{\normalsize\bfseries}{\thesubsection}{0.65em}{}
\titlespacing*{\section}{0pt}{1.4em}{0.45em}
\titlespacing*{\subsection}{0pt}{0.9em}{0.25em}
\newcolumntype{Y}{>{\raggedright\arraybackslash}X}

\begin{document}

\begin{center}
{\Large\bfseries GaitFM: A Cross-Modal Foundation Model for Wearable Gait Assessment}\par
\vspace{0.5em}
{\large Yize Cai$^{1}$, Zhiqing Hong$^{1,2,\dagger}$}\par
\vspace{0.35em}
{\small $^{1}$The Hong Kong University of Science and Technology (Guangzhou), Guangzhou, China.\\
\small $^{2}$The Hong Kong University of Science and Technology, Hong Kong, China.\\
$^{\dagger}$Corresponding author: zhiqinghong@hkust-gz.edu.cn}\par

\end{center}

\begin{abstract} 
Gait assessment supports the diagnosis, monitoring, and prognosis of neurological and musculoskeletal disorders.
Wearable inertial measurement units (IMUs) offer an accessible and low-cost approach to automated gait assessment. 
However, IMU measurements vary substantially across populations, body locations, and device configurations, while existing deep learning models are often developed using small, task- and configuration-specific datasets, limiting their generalizability across diverse clinical conditions and real-world settings.
Here we present GaitFM, a cross-modal foundation model for IMU-based gait assessment that transfers knowledge from pretrained large vision models and learns from large-scale unlabeled sensor data.
We pretrain and evaluate GaitFM on a large curated dataset comprising about 4,000 participants and 1.4 million gait windows across diverse clinical and daily conditions, sensor modalities, and body locations. 
Across 17 gait-related tasks, including three gait parameter estimation tasks, 13 disease and abnormality diagnosis tasks, and one clinical outcome prediction task, GaitFM consistently benefits from visual pretraining and outperforms existing end-to-end models. 
GaitFM generalizes across sensor configurations and body locations used in clinical systems and everyday wearable devices and supports low-latency on-device inference. 
We further demonstrate two applications of GaitFM: zero-shot extraction of gait features that provide predictive information beyond demographic covariates and retrieval of clinically similar cases to support clinician review.
Our analyses show that visual knowledge transfer helps GaitFM capture local gait patterns and intersensor-modality dependencies, improves downstream data efficiency, and produces increasing gains as the visual backbone scales. 
Together, these findings establish visual knowledge transfer as an effective strategy for mitigating data scarcity in wearable sensing and clinical applications.

\end{abstract}


\section*{Introduction}

Neurological and musculoskeletal disorders pose a growing global health burden. 
Current projections indicate that Parkinson’s disease (a neurological disorder) will affect more than 14 million people worldwide by 2040~\cite{tabar2025phase}, while the number of osteoarthritis cases (a musculoskeletal disorder) is expected to increase by 49–95\% by 2050~\cite{steinmetz2023global}.
Among individuals with these conditions, alterations in the characteristic pattern of movement during walking, i.e., gait impairment, are a common clinical manifestation and can be among their most disabling and distressing features\cite{mirelman2019gait, thenaisie2022principles, buckley2019role, winters1987gait, mills2013biomechanical, kaufman2001gait, hart2016knee, de2022evaluation, balachandar2026neuromodulation}.
In addition, evidence indicates that gait patterns evolve with symptom severity~\cite{liu2022monitoring, galna2015progression} and can provide information about future clinical risks~\cite{studenski2011gait, verghese2009quantitative}.
Therefore, quantitative gait assessment, which objectively characterizes fine-grained walking patterns, has become an important clinical tool for detecting and evaluating disorders~\cite{mills2013biomechanical, kaufman2001gait, hart2016knee, de2022evaluation, mirelman2019gait, thenaisie2022principles, buckley2019role, winters1987gait}, monitoring disease progression~\cite{olney1996hemiparetic, salomon2024machine}, and predicting adverse clinical outcomes~\cite{maki1997gait, kasper2025wearable}.

Conventional gait assessment typically relies on either observational clinical scales or instrumented laboratory analysis. 
In the former, patients perform standardized tasks in a hospital or clinic, and trained assessors visually evaluate and score their movements~\cite{goetz2008movement, schmitz2006scale, hulzinga2020new}. 
Although readily accessible, these assessments are inherently subjective and can vary across observers~\cite{albanese1994ensuring, goetz2004assuring, subramony2005measuring}. 
Instrumented gait analysis provides more detailed and objective measurements of gait kinematics and kinetics, but generally requires specialized facilities, expensive systems such as optical motion-capture platforms and trained personnel~\cite{tan2026gaitdynamics, yamada2025utility}. 
These examinations are time-consuming and can be particularly burdensome for individuals with limited mobility. 
More importantly, both approaches capture only brief snapshots of gait under controlled conditions and can therefore miss subtle, intermittent or context-dependent abnormalities that emerge during daily life~\cite{kasper2025wearable}. 

Data-driven deep learning (DL) models offer a path towards automated gait assessment by learning clinically relevant patterns directly from gait-related data, thereby reducing reliance on manual evaluation and complex analytical pipelines. 
Existing DL-based methods have leveraged sensing modalities such as vision~\cite{kidzinski2020deep, yamada2025utility} and floor sensors~\cite{shi2020deep} to characterize gait patterns.
However, vision-based measurements can be sensitive to environmental conditions~\cite{migueles2017accelerometer}, whereas floor-based systems restrict assessment to instrumented spaces.
Wearable sensors, such as force sensors~\cite{liu2026unveiling}, electromyography (EMG)~\cite{kim2026simplified}, and inertial measurement units (IMUs)~\cite{nguyen2023accelerometer, kasper2025wearable, kadirvelu2023wearable}, offer a means for continuous gait monitoring in daily life~\cite{schalkamp2023wearable, brognara2019assessing}. 
Among these modalities, IMUs are particularly suitable for scalable gait assessment because they are widely integrated into everyday consumer devices and can capture fine-grained body movement at low cost~\cite{bai2014quantitative, kasper2025wearable}.
However, existing IMU-based gait studies typically rely on heterogeneous datasets that vary in sensing devices, body locations, sensor combinations, and acquisition protocols~\cite{kasper2025wearable, kadirvelu2023wearable, salomon2024machine}.
As a result, models developed for specific sensor configurations and clinical tasks often exhibit limited generalizability when deployed in new settings~\cite{yamada2025utility}.
Realizing the potential of IMU-based gait assessment therefore requires a unified framework that can learn generalizable representations across diverse sensor configurations and gait-related tasks.

Foundation models offer a promising solution to addressing this challenge by learning robust and generalizable representations through pretraining on large-scale datasets, which can then be reused across diverse sensor configurations, body locations, and downstream tasks. 
The success of this paradigm has been demonstrated by large language models (LLMs) and large vision models (LVMs)~\cite{bommasani2021opportunities}, with growing applications in medical domains such as pathology~\cite{xu2024whole, chen2024towards, wang2024pathology}, radiology~\cite{wu2025towards}, spatial proteomics~\cite{wenckstern2026virtual}, and sleep medicine~\cite{thapa2026multimodal}.
However, its application to IMU-based gait assessment is constrained by the scarcity of sufficiently large and diverse gait data for pretraining.
This scarcity is partly attributable to the sensitive nature of IMU sensor data, as stringent privacy regulations and requirements for protecting health information complicate data collection, management and sharing~\cite{tudosiu2024realistic, yamada2025utility}.
Cross-modal knowledge transfer offers a potential means of overcoming this limitation by adapting representations learned from data-rich to data-scarce modalities. 
Previous studies have shown that this approach can alleviate data scarcity and improve downstream performance~\cite{qiu2026large, zhuang2025advancing, he2024mosaic, sun2025automated}. 
Motivated by these advances, we investigate the potential of transferring knowledge learned from natural images to IMU-based gait assessment tasks.

Here we introduce GaitFM, a Transformer-based gait foundation model that leverages pretrained visual knowledge to map heterogeneous gait-related IMU data into a shared and unified representation space, enabling a single model to operate across sensor configurations, body locations, and tasks. 
To maximize representation learning from limited sensor data, we curate a dataset comprising approximately 4,000 participants and 1.41 million gait windows. 
GaitFM transfers knowledge across modalities by initializing a masked autoencoder (MAE) architecture with weights pretrained on over one million natural images, followed by continued masked pretraining on the curated sensor dataset.

Across tasks including gait parameter estimation, disease and abnormality diagnosis, and clinical outcome prediction, GaitFM achieves strong performance, demonstrating broad cross-task generalizability. 
Furthermore, GaitFM accommodates dynamic sensor inputs without retraining, operates across clinical and everyday settings, and supports on-device deployment on consumer devices, demonstrating strong generalizability across deployment settings and the potential for low-cost gait assessment in daily life.
Beyond task performance, GaitFM shows broader application potential. 
The pretrained GaitFM enables zero-shot extraction of gait features beyond basic demographic characteristics, thereby providing additional clinically relevant information for clinical analysis.
In Parkinson’s disease, similar gait representations extracted by pretrained GaitFM correspond to greater clinical similarity between cases, enabling the retrieval of comparable cases to support clinician review.
Analyses of visual knowledge transfer further show that visual representations provide a powerful source of transferable knowledge to mitigate gait-related sensor data scarcity, support learning local gait patterns and relationships between sensor modalities, improve downstream data efficiency, and enable sensor representation learning to benefit from scaling behaviors in the vision domain. 

\begin{figure}[t]
	\centering
	\includegraphics[width=6.9in]{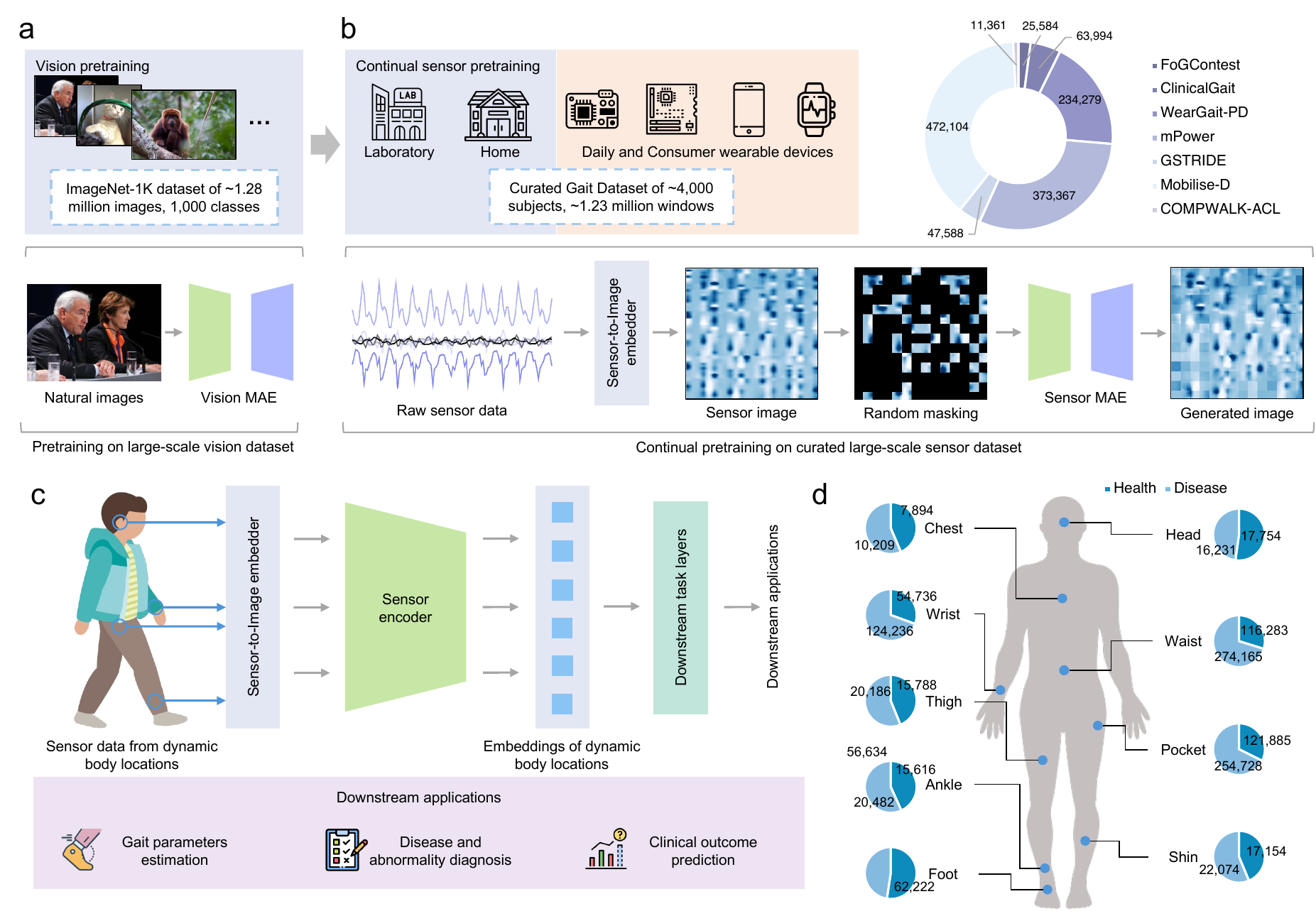}
	\caption{\textbf{Overview of the GaitFM framework.}
    \textbf{a,} GaitFM adopts a masked autoencoder (MAE) architecture initialized with weights pretrained on ImageNet-1K dataset~\cite{deng2009imagenet}, comprising approximately 1.28 million natural images across 1,000 classes.
    \textbf{b,} A sensor-to-image embedder enables continued MAE pretraining and evaluation on a curated gait dataset comprising approximately 4,000 participants and 1.41 million IMU gait windows from seven heterogeneous datasets.
    \textbf{c,} Fine-tuning of GaitFM for downstream gait assessment tasks.
    GaitFM accommodates varying sensor placements and modality combinations, supporting flexible gait assessment across clinical and daily settings.
    The downstream tasks encompass gait parameter estimation, disease and abnormality diagnosis, and clinical outcome prediction.
    \textbf{d,} Distribution of samples across sensor placements.} 
	\label{fig:framework}
\end{figure}

\section*{Results}

\subsection*{A foundation model for gait monitoring}

GaitFM employs a cross-modality transfer learning framework designed to leverage pretrained knowledge from the data-rich visual domain to alleviate data scarcity in IMU sensing and learn generalizable gait representations.
We initialize the masked autoencoder (MAE) architecture with weights pretrained via masked image reconstruction on the large-scale ImageNet-1K dataset~\cite{deng2009imagenet}, which contains about 1.28 million natural images spanning 1,000 classes (Fig.~\ref{fig:framework}a). 
To bridge the structural differences between images and sensor data, we introduce a sensor-to-image embedder that transforms raw IMU data into image-like representations compatible with the pretrained vision encoder (Fig.~\ref{fig:framework}b).
We then continue pretraining the model through masked sensor reconstruction on a curated gait IMU dataset comprising about 4,000 participants and 1.41 million gait windows recorded across diverse body locations (Fig.~\ref{fig:framework}d).
Detailed dataset descriptions are provided in Supplementary Information Section 1.

For downstream tasks, the sensor-to-image embedder and the pretrained MAE encoder are retained, whereas the decoder is discarded and replaced with task-specific layers (Fig.~\ref{fig:framework}c). 
For gait parameter estimation tasks, the latent embeddings produced by the pretrained encoder are processed by a residual network (ResNet)-based prediction head before regression. 
For disease and abnormality diagnosis and clinical outcome prediction tasks, the encoder embeddings are fed into lightweight multilayer perceptron (MLP) heads for classification or prediction. 
The sensor-to-image embedder and MAE encoder independently process each body-location input to generate latent embeddings, which are subsequently aggregated and passed to the downstream layers. 
We randomly sample subsets of body locations and sensor modalities as model inputs during fine-tuning, allowing GaitFM to generalize across dynamic sensor combinations and input configurations without retraining.

\subsection*{GaitFM generalizes across diverse healthcare tasks}

\begin{figure*}[htbp]
    \centering
    \includegraphics[width=\textwidth]{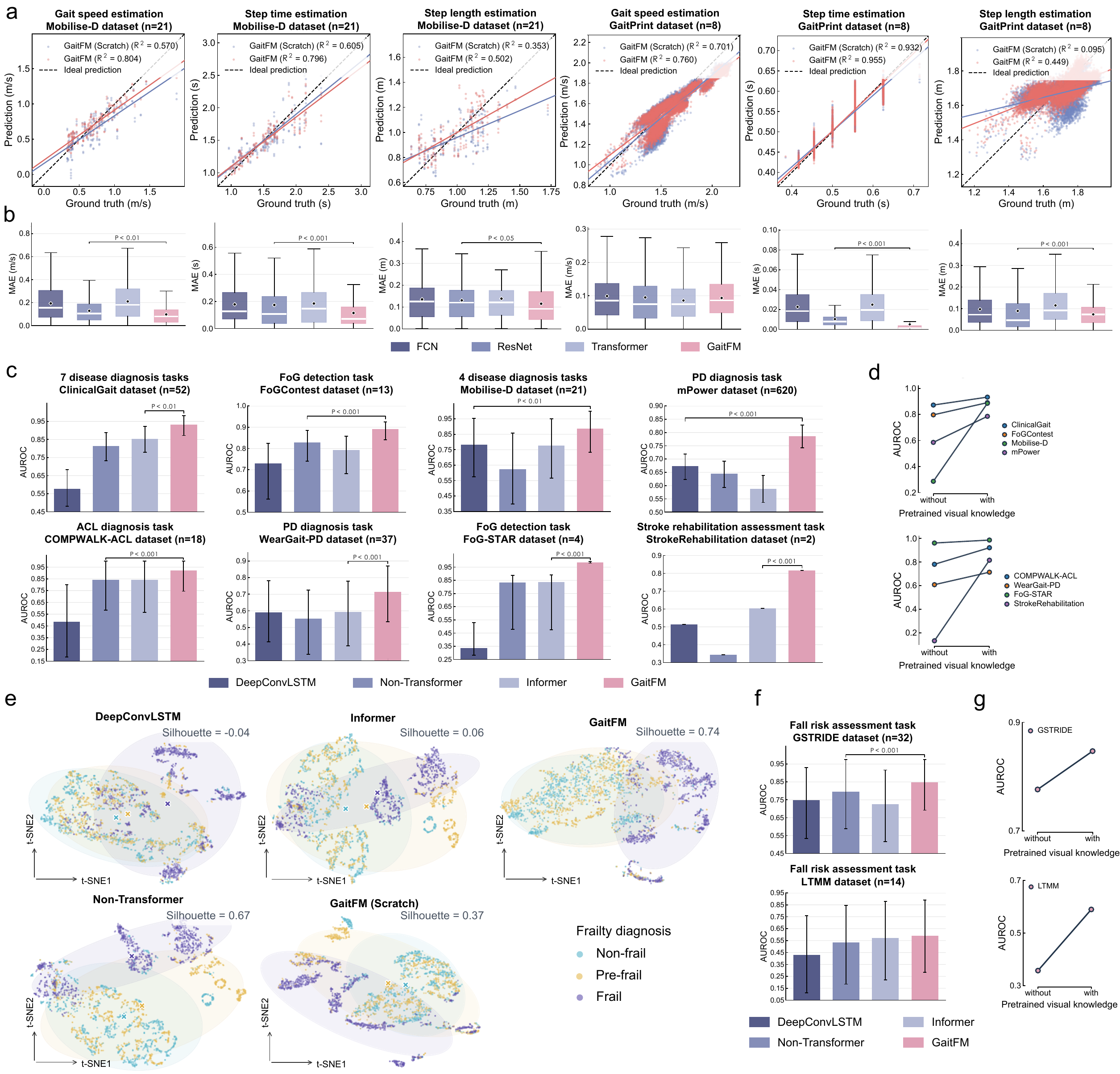}
    \caption{\textbf{GaitFM achieves strong performance across diverse gait assessment tasks and benefits from pretrained visual knowledge.}
    \textbf{a,b,} Comparison of gait parameter estimation tasks on the internal Mobilise-D dataset and the external GaitPrint dataset.
    \textbf{a,} Coefficient of determination ($R^2$) comparison for identical model architectures with and without visual pretraining. 
    \textbf{b,} Mean absolute error (MAE) comparison across three gait parameter estimation tasks.
    \textbf{c,d,} Comparison of disease and gait abnormality diagnosis tasks on six internal datasets (ClinicalGait, FoGContest, Mobilise-D, mPower, COMPWALK-ACL and WearGait-PD) and two external datasets (FoG-STAR and StrokeRehabilitation).
    \textbf{c,} Area under the receiver operating characteristic curve (AUROC) comparison across disease and gait abnormality diagnosis tasks.
    \textbf{d,} AUROC comparison of identical model architectures with and without visual pretraining.
    \textbf{e,} t-SNE visualization of learned representations for frailty diagnosis in the GSTRIDE dataset.
    GaitFM achieves a higher silhouette score than end-to-end models and its counterpart without visual pretraining, indicating improved separation between classes.
    \textbf{f,g,} Comparison of clinical outcome prediction task on the internal GSTRIDE dataset and the external LTMM dataset.
    \textbf{f,} AUROC comparison on clinical outcome prediction task.
    \textbf{g,} AUROC comparison of identical model architectures with and without visual pretraining. 
    Error bars denote 95\% confidence intervals (CIs) obtained from 1,000 bootstrap resamples.
    $P$ values are calculated using paired bootstrapping for gait parameter estimation and two-sided Wilcoxon signed-rank tests for disease and gait abnormality diagnosis and clinical outcome prediction.}
    \label{fig:results1}
\end{figure*}

We systematically benchmark GaitFM on 17 downstream tasks across seven internal datasets (included in pretraining) and four external datasets (not included in pretraining). 
These tasks span three application domains: gait parameter estimation (including gait speed, step length, and step time estimation),  disease and abnormality diagnosis (including disease classification and grading, episode detection, and progression assessment), and clinical outcome prediction (including fall risk assessment). 
Gait parameter estimation is formulated as a regression task, whereas disease and abnormality diagnosis and clinical outcome prediction are formulated as a classification task.
Detailed task descriptions are provided in Supplementary Information Section 3.1.

Across all tasks, we compare GaitFM with GaitFM (Scratch), an architecture-matched MAE model that undergoes the same sensor pretraining but is randomly initialized rather than from vision pretrained weights. 
This comparison isolates the contribution of cross-modal vision knowledge transfer. 
For regression, we additionally compare GaitFM with end-to-end fully convolutional network (FCN)~\cite{wang2017time}, ResNet~\cite{wang2017time} and Transformer~\cite{katrompas2022recurrence}. 
For classification, the baseline models include DeepConvLSTM~\cite{ordonez2016deep}, Informer~\cite{zhou2021informer}, and Non-stationary Transformer~\cite{liu2022nonstationary}.
Detailed descriptions of end-to-end baselines are provided in Supplementary Information Section 2.1.

\textbf{Gait parameter estimation.}
We evaluate gait parameter estimation tasks on the internal Mobilise-D dataset~\cite{mazza2021technical} and the external GaitPrint dataset~\cite{wiles2025nonan}, using gait speed, step length, and step time as regression targets. 
Mobilise-D provides accelerometer and gyroscope data from a sensor worn on the lower back, whereas GaitPrint provides accelerometer and gyroscope data from five locations: the head, both wrists and both thighs.

Compared with GaitFM (Scratch), GaitFM increases $R^2$ by 0.234 ($P<0.05$), 0.149, and 0.191 ($P<0.01$) for the three respective targets on Mobilise-D (Fig.~\ref{fig:results1}a). 
The corresponding improvements on GaitPrint are 0.059, 0.354, and 0.023 ($P<0.01$).
These gains indicate that visual knowledge can be effectively transferred to improve gait parameter estimation tasks.
GaitFM also outperforms the conventional end-to-end regression models (Fig.~\ref{fig:results1}b).
Compared with the best-performing conventional baseline, it reduces MAE by 0.031 ($P<0.01$), 0.059 ($P<0.001$) and 0.016 ($P<0.05$) for gait speed, step time and step length, respectively, on Mobilise-D. 
In addition, GaitFM maintains strong performance on the external GaitPrint dataset, highlighting its ability to generalize gait parameter estimation across independent cohorts.

\textbf{Disease and abnormality diagnosis.}
We evaluate GaitFM on six internal datasets (ClinicalGait~\cite{voisard2025dataset}, Mobilise-D~\cite{mazza2021technical}, mPower~\cite{bot2016mpower}, COMPWALK-ACL~\cite{yona2025compwalk}, FoGContest~\cite{salomon2024machine} and WearGait-PD~\cite{anderson2024weargait}) and two external datasets (FoG-STAR~\cite{borzi2026multi} and StrokeRehabilitation~\cite{zhou_2024_10534055}).
For neurological and musculoskeletal disease diagnosis, ClinicalGait provided accelerometer and gyroscope measurements from the head, waist and both feet for seven tasks: hip osteoarthritis (HOA), knee osteoarthritis (KOA), anterior cruciate ligament (ACL) injury, cerebrovascular accident (CVA), Parkinson’s disease (PD), chemotherapy-induced peripheral neuropathy (CIPN) and radiation-induced leukoencephalopathy (RIL).
Mobilise-D provided lower-back accelerometer and gyroscope measurements for four tasks: PD, multiple sclerosis (MS), chronic obstructive pulmonary disease (COPD) and congestive heart failure (CHF).
mPower provides accelerometer and gyroscope measurements from a smartphone carried in the trouser pocket for PD diagnosis.
COMPWALK-ACL contained accelerometer measurements from seven lower-body locations, comprising the waist, both thighs, both shins and both feet, for ACL diagnosis.
WearGait-PD provided accelerometer and gyroscope measurements from the head, both wrists and both thighs for PD diagnosis.
For disease episode detection, we evaluate freezing-of-gait (FoG) detection using back-mounted accelerometer data from FoGContest and accelerometer and gyroscope data from the wrist, lower back and both ankles in FoG-STAR. 
For post-disease recovery assessment, accelerometer data collected from both feet, both wrists and the sacral region in the StrokeRehabilitation dataset were used to evaluate recovery following stroke.

The comparison results of GaitFM and baselines are shown in Fig.~\ref{fig:results1}c.
For disease diagnosis, GaitFM improves average area under the receiver operating characteristic curve (AUROC) over the best-performing conventional baseline by 0.078 ($P<0.01$) on ClinicalGait and 0.104 ($P<0.01$) on Mobilise-D, with results averaged across seven and four diagnostic tasks, respectively. 
It further improves AUROC by 0.112 ($P<0.001$) on mPower, 0.078 ($P<0.001$) on COMPWALK-ACL and 0.121 ($P<0.001$) on WearGait-PD. 
For episode detection of FoG, GaitFM improves AUROC by 0.063 ($P<0.001$) on FoGContest and 0.149 ($P<0.001$) on the external FoG-STAR dataset. 
For recovery assessment, we evaluate post-stroke functional recovery on the external StrokeRehabilitation dataset. 
GaitFM achieves an AUROC of 0.816, outperforming the best-performing conventional baseline by 0.214 ($P<0.001$).
We then examine the contribution of vision-pretrained weights to disease and abnormality diagnosis. 
Across all eight datasets, cross-modal visual knowledge increases AUROC by 0.025-0.683 ($P<0.001$) compared with GaitFM (Scratch) (Fig.~\ref{fig:results1}d).
These consistent improvements demonstrate the effectiveness of transferring visual knowledge to disease and abnormality diagnosis tasks.

To assess representation quality, we visualize the embeddings learned for frailty diagnosis on GSTRIDE using t-distributed stochastic neighbour embedding (t-SNE, Fig.~\ref{fig:results1}e). 
GaitFM achieves the highest silhouette coefficient (0.74 versus 0.67 for the second-best model), indicating better separation between classes in the learned representation space.
Moreover, compared with GaitFM (Scratch), GaitFM also yields clearer separation between normal and abnormal gait (0.74 versus 0.37), indicating that visual knowledge improves their discriminability.
Detailed disease and abnormality diagnosis results are provided in Supplementary Information Section 4.

\textbf{Clinical outcome prediction.}
We use the fall risk assessment task to evaluate GaitFM for clinical outcome prediction. 
Unlike the cross-sectional disease and abnormality diagnosis tasks, fall risk assessment is longitudinal and assesses clinical outcomes.
In the internal GSTRIDE dataset, foot-mounted accelerometer and gyroscope measurements were used to predict falls within one year. 
In the external LTMM dataset, one-year fall risk was predicted from lower-back accelerometer measurements.

GaitFM achieves the highest AUROC in both settings, outperforming the best-performing conventional end-to-end baseline by 0.051 on GSTRIDE ($P<0.001$) and 0.018 on LTMM (Fig.~\ref{fig:results1}f). 
Vision-pretrained weights increase AUROC from 0.777 to 0.847 ($P<0.001$) on GSTRIDE and from 0.357 to 0.589 on LTMM, corresponding to absolute gains of 0.071 and 0.232, respectively (Fig.~\ref{fig:results1}g). 
These gains indicate that visual knowledge can be effectively transferred to improve performance on the clinical outcome prediction task.

\subsection*{Knowledge transfer from vision models enhances gait-related sensor representation learning}

\begin{figure*}[htbp]
    \centering
    \includegraphics[width=\textwidth]{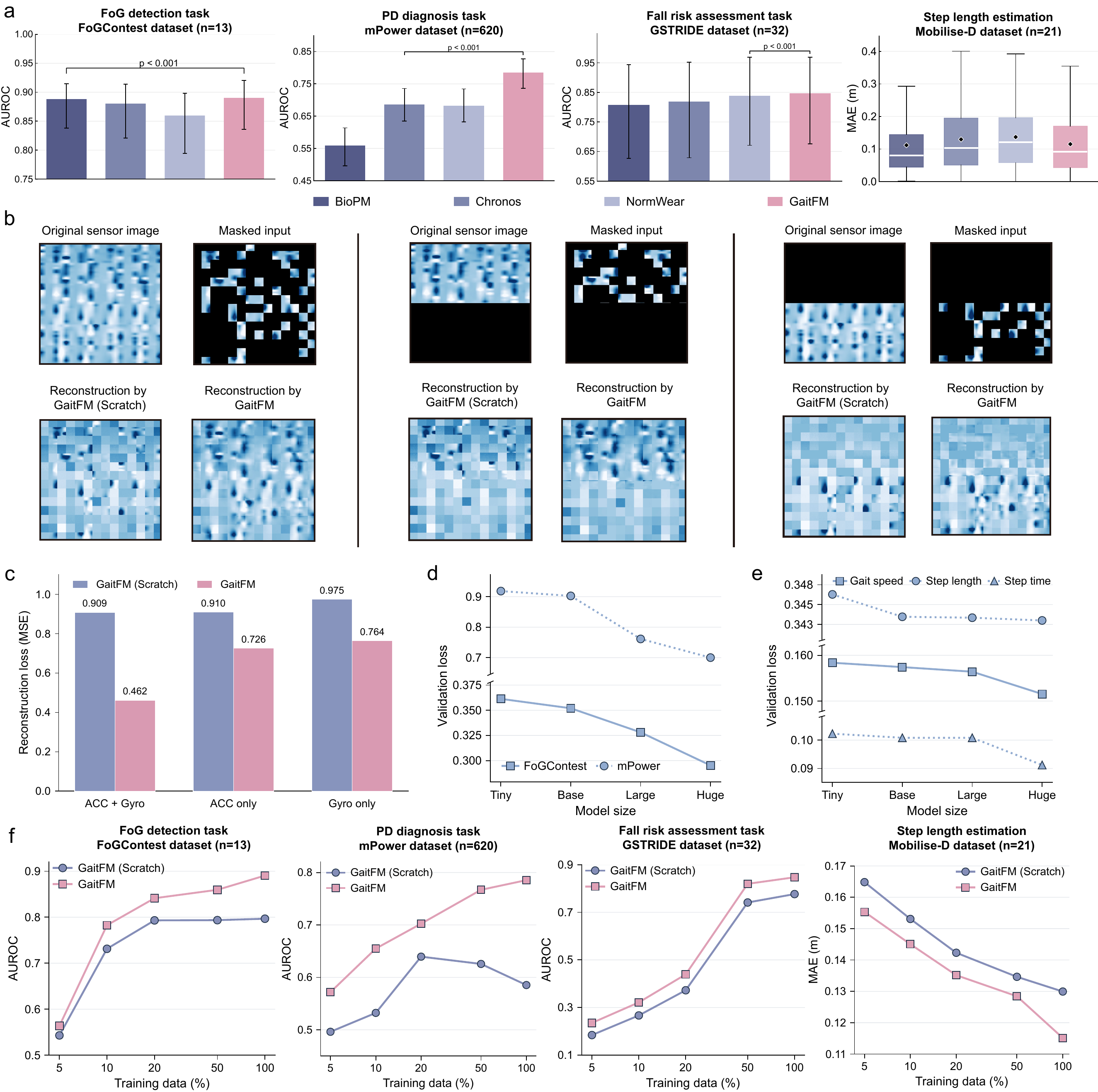}
    \caption{\textbf{Knowledge transfer from vision models enhances gait-related sensor representation learning.}
    \textbf{a,} Performance comparison of GaitFM with foundation models pretrained on other modalities: general time-series data (Chronos~\cite{ansari2024chronos}), wearable physiological data (NormWear~\cite{luo2026toward}), and daily-life IMU data (BioPM~\cite{tarale2026bioinspired}). 
    Each model was initialized with its original pretrained weights and then further pretrained on the same gait training set used for GaitFM.
    GaitFM achieves performance comparable to or better than that of these models, demonstrating the transferability of visual knowledge to sensor-data modeling.
    \textbf{b,} Representative examples of masked sensor image reconstruction with accelerometer and gyroscope data (acc + gyro, left), accelerometer data only (acc only, middle), and gyroscope data only (gyro only, right).
    \textbf{c,} Reconstruction loss of GaitFM with and without visual knowledge (Scratch). 
    Visual knowledge reduces reconstruction loss across all three input settings, suggesting that it helps GaitFM capture local sensor patterns and cross-modality correlations.
    \textbf{d,e,} Effect of model scale on classification and regression tasks, evaluated using the Tiny, Base, Large, and Huge variants of the LVM on FoGContest, mPower (\textbf{d}) and Mobilise-D (\textbf{e}).
    \textbf{f,} Data efficiency of GaitFM and GaitFM (Scratch) when fine-tuned with 5\%, 10\%, 20\%, 50\%, and 100\% of the downstream training data. 
    GaitFM with visual knowledge consistently achieves better performance, demonstrating improved data efficiency through visual knowledge transfer.}
    \label{fig:results3}
\end{figure*}

Traditional foundation models typically require large-scale datasets to learn robust and generalizable representations that can adapt to diverse scenarios and downstream tasks.
However, this requirement is difficult to apply to IMU-based gait assessment because large-scale gait data remain scarce. 
We demonstrate that visual pretraining provides a powerful source of transferable knowledge that substantially improves sensor data understanding, thereby reducing the model’s dependence on large-scale sensor datasets and labeled downstream data.

\textbf{Visual knowledge provides a powerful source of modality knowledge for mitigating sensor-data scarcity.} 
To evaluate the effectiveness of visual knowledge as a source for IMU-based gait assessment tasks, we compared GaitFM with foundation models pretrained on other modalities: (1) BioPM~\cite{tarale2026bioinspired}, pretrained on daily-life IMU data, (2) Chronos~\cite{ansari2024chronos}, pretrained on general time-series data, and (3) NormWear~\cite{luo2026toward}, pretrained on physiological data. 
Detailed descriptions of foundation models are provided in Supplementary Information Section 2.2.
Each model was initialized with its original pretrained weights and further pretrained on the same gait training set as GaitFM, following its respective pretraining strategy. 
As shown in Fig.~\ref{fig:results3}a, we evaluated the models on four downstream tasks: FoG episode detection (FoGContest), PD diagnosis (mPower), fall risk assessment (GSTRIDE), and step length estimation (Mobilise-D). 
GaitFM achieves performance comparable to or better than that of these foundation models across tasks, supporting the effectiveness of visual knowledge as a source of transferable knowledge for IMU-based gait assessment.
Detailed results are provided in Supplementary Information Section 4.2.

\textbf{Visual knowledge helps model capture local gait patterns and cross-modal relationships.}
To evaluate how visual knowledge affects GaitFM’s ability to understand sensor data, we evaluated masked sensor image reconstruction under three input settings: accelerometer and gyroscope data (acc + gyro, Fig.~\ref{fig:results3}b left), accelerometer data only (acc only, Fig.~\ref{fig:results3}b middle), and gyroscope data only (gyro only, Fig.~\ref{fig:results3}b right). 
The reconstruction MSE is shown in Fig.~\ref{fig:results3}c. 
In the acc + gyro setting, GaitFM reduces the MSE from 0.909 to 0.462 compared with GaitFM (Scratch), suggesting that visual knowledge helps the model capture local patterns in sensor data. 
In the acc-only and gyro-only settings, the MSE decreases from 0.910 to 0.726 and from 0.975 to 0.764, respectively. 
These reductions suggest that visual knowledge also helps GaitFM capture intersensor-modality relationships.

\textbf{Gait assessment tasks benefit from the scaling behavior of LVMs.}
To examine how the scale of the visual model affects its ability to model IMU sensor data, we evaluated GaitFM variants built on Tiny, Base, Large, and Huge LVM backbones. 
As shown in Fig.~\ref{fig:results3}d, validation loss decreases with increasing model size for both FoG episode detection (FoGContest) and PD diagnosis (mPower). 
A similar scaling trend is observed for gait parameter estimation (Mobilise-D), with larger backbones achieving lower validation loss for gait speed, step length and step time estimation tasks (Fig.~\ref{fig:results3}e).

\textbf{Visual knowledge improves downstream data efficiency of gait assessment tasks.}
To evaluate how visual knowledge affects performance when labeled downstream data are limited, we fine-tuned GaitFM and GaitFM (Scratch) using 5\%, 10\%, 20\%, 50\%, and 100\% of the training data for each task. 
As shown in Fig.~\ref{fig:results3}f, GaitFM with visual knowledge consistently outperforms GaitFM (Scratch) across FoG episode detection (FoGContest), PD diagnosis (mPower), fall risk assessment (GSTRIDE), and step length estimation (Mobilise-D). 
These results demonstrate that visual knowledge improves the data efficiency of IMU-based gait assessment.

\subsection*{GaitFM generalizes across dynamic sensor placements and configurations}

Sensor configurations vary substantially in their number, anatomical placement and sensing modality across real-world settings. 
A robust gait foundation model should therefore accommodate this variability without retraining or architectural modification~\cite{gu2026cardiac, bian2026improving}. 
We evaluate GaitFM across dynamic combinations of sensor locations within a unified framework. 
We further examine the feasibility of accessible, low-cost gait assessment using sensor placements corresponding to everyday wearable devices.

\begin{figure*}[t]
	\centering
    \includegraphics[width=\textwidth]{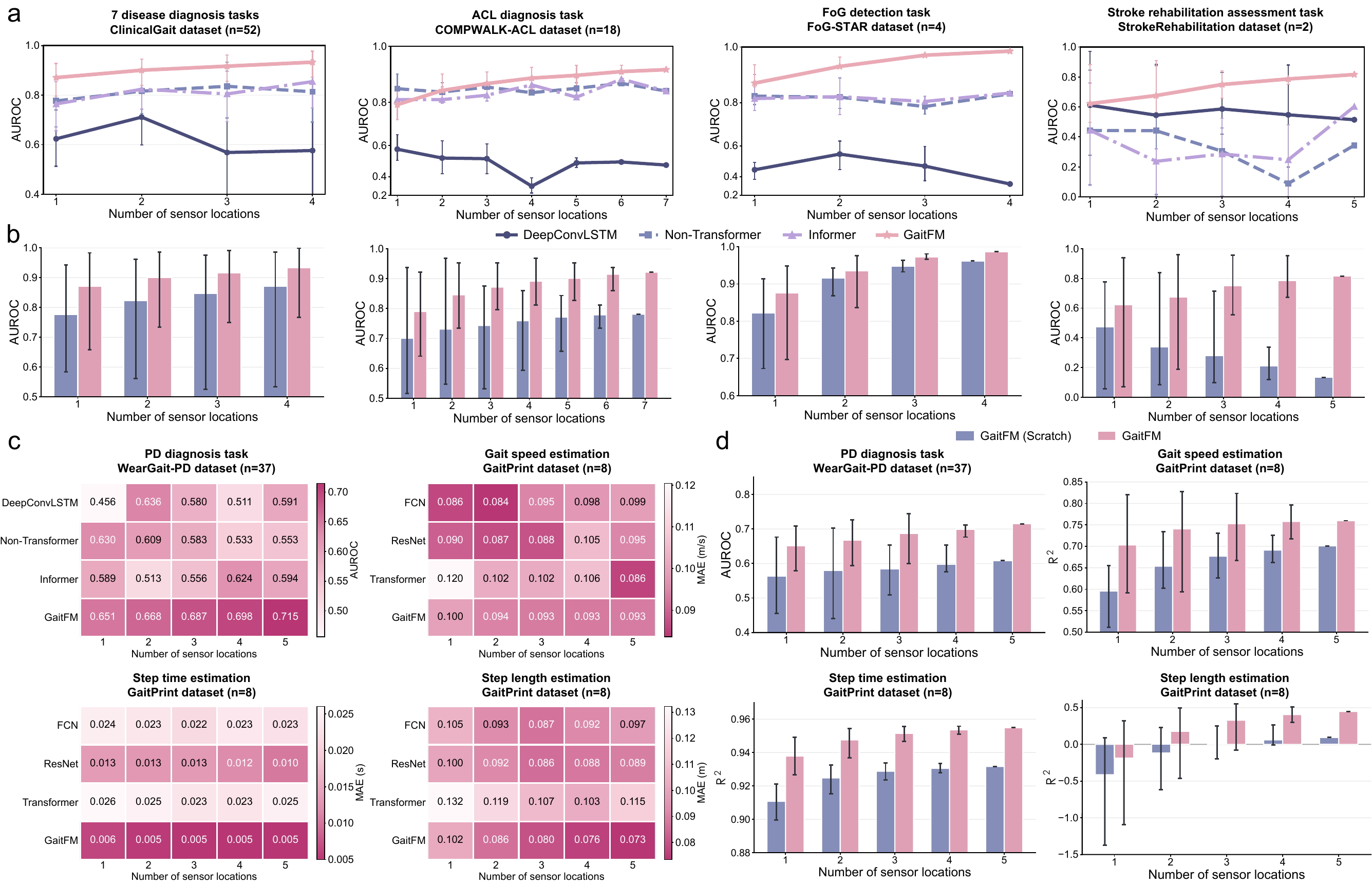}
	\caption{\textbf{GaitFM generalizes across dynamic sensor placements and configurations.}
	\textbf{a,} Performance on disease and abnormality diagnosis tasks across dynamic combinations of dedicated sensor locations in ClinicalGait, COMPWALK-ACL, FoG-STAR and StrokeRehabilitation.
	\textbf{b,} Contribution of vision-pretrained knowledge under dynamic combinations of dedicated sensor locations.
	\textbf{c,} Performance on PD diagnosis and gait parameter estimation in WearGait-PD and GaitPrint across different combinations of simulated sensor placements representative of everyday wearable devices (left and right trouser pockets, left and right wrists, and head).
	\textbf{d,} Contribution of vision-pretrained knowledge under dynamic combinations of everyday sensor locations.}
	\label{fig:results2}
\end{figure*}

\textbf{Performance under dedicated sensor location settings.}
We first evaluate the ability of GaitFM to accommodate varying numbers and combinations of dedicated gait-analysis IMU sensors. 
The evaluation comprises ClinicalGait (7 diagnostic tasks and dynamic combinations of 4 sensor locations), COMPWALK-ACL (ACL diagnosis and combinations of 7 lower-limb locations), FoG-STAR (FoG detection and combinations of 4 locations) and StrokeRehabilitation (post-stroke recovery assessment and combinations of 5 locations). 
ClinicalGait and FoG-STAR provide both accelerometer and gyroscope measurements, whereas COMPWALK-ACL and StrokeRehabilitation provide accelerometer measurements only. 
Across these datasets, GaitFM outperforms conventional end-to-end models under varying numbers and configurations of sensor locations (Fig.~\ref{fig:results2}a). 
Performance generally improves as additional locations are incorporated, indicating that GaitFM effectively integrates complementary information across body locations without retraining or architectural modification.
Furthermore, as shown in Fig.~\ref{fig:results2}b, vision-pretrained knowledge consistently improves performance across all sensor-location combinations, demonstrating its effective transfer to diverse dedicated sensor configurations.

\textbf{Performance under daily sensor location settings.}
We further investigate the potential of IMU data collected from locations corresponding to everyday wearable devices for gait analysis. 
We select five body locations from WearGait-PD and GaitPrint: bilateral wrist recordings serve as proxies for smartwatches, bilateral upper-thigh recordings as proxies for smartphones carried in trouser pockets, and head-mounted recordings as proxies for ear-worn devices or smart glasses. 
GaitFM maintains strong performance across these sensor configurations for PD diagnosis on WearGait-PD and gait parameter estimation on GaitPrint (Fig.~\ref{fig:results2}c), supporting the feasibility of gait assessment using everyday consumer devices. 
Moreover, vision-pretrained knowledge consistently improves performance across all combinations of daily sensor locations (Fig.~\ref{fig:results2}d), demonstrating that cross-modal visual knowledge remains transferable across daily sensor configurations.
Detailed results for all sensor configurations are provided in Supplementary Information Section 4.

\subsection*{Clinical applications of GaitFM}

Clinical gait analysis has advanced substantially in recent years, while handcrafted and configuration-specific features remain widely used in this area.
However, it remains unclear how well these features generalize across recording settings and sensor configurations.
We show that a frozen GaitFM encoder extracts informative gait representations that support downstream clinical applications.

\begin{figure*}[ht]
    \centering
    \includegraphics[width=6.7in]{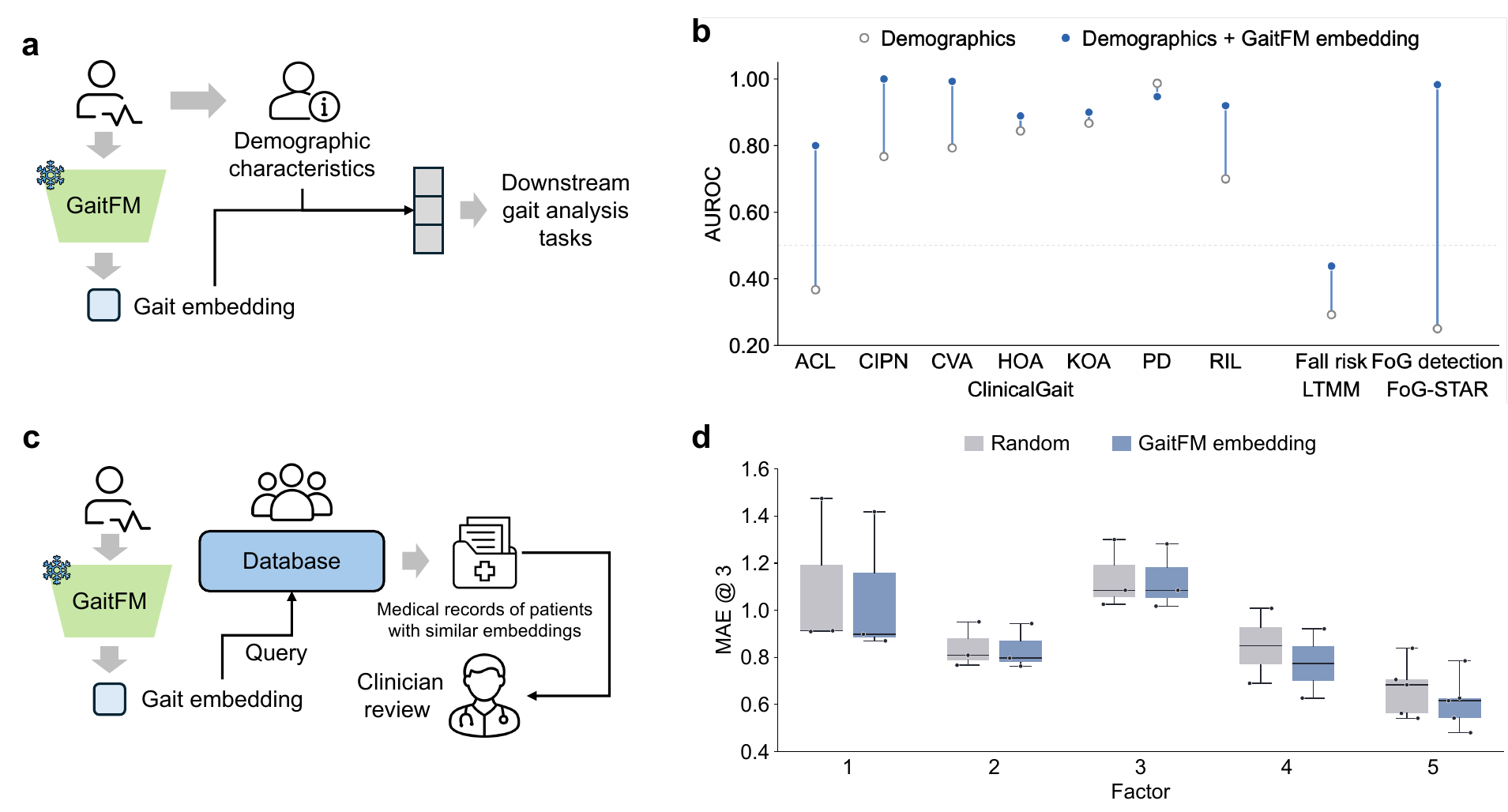}
    \caption{\textbf{Clinical applications of GaitFM.}
    \textbf{a,} Pretrained GaitFM extracts gait representations that complement demographic covariates, providing reusable features for clinical assessment.
    \textbf{b,} Comparison of models fitted using demographic covariates alone with those incorporating both demographic covariates and GaitFM embeddings across one internal dataset (ClinicalGait) and two external datasets (LTMM and FoG-STAR). 
    Adding GaitFM embeddings improves AUROC in eight of nine tasks, indicating that the embeddings capture predictive gait information beyond demographic covariates.
    \textbf{c,} Pretrained GaitFM retrieves clinically similar patients by computing the cosine distance between a query embedding and embeddings linked to clinical records in a reference database, enabling the retrieved records to be reviewed by clinicians.
    \textbf{d,} Comparison of GaitFM-based retrieval and random selection among patients with PD in mPower. 
    Each box shows the distribution of average absolute score differences among the three retrieved cases (MAE@3) for one of five Movement Disorder Society-Unified Parkinson’s Disease Rating Scale (MDS-UPDRS)~\cite{goetz2008movement} factors, with lower values indicating greater clinical similarity.}
    \label{fig:application}
\end{figure*}

\textbf{Zero-shot extraction of reusable features beyond demographic characteristics.}
Demographic covariates such as age, gender, and body mass index (BMI) are associated with a range of health outcomes and serve as readily available predictors for gait-related tasks~\cite{gabet2026gait, lopez2023hallmarks}.
We show that gait embeddings extracted from IMU recordings by the pretrained GaitFM add predictive information beyond these covariates, supporting their potential use in clinical gait assessment and diagnosis (Fig.~\ref{fig:application}a).
Using a frozen GaitFM encoder, we extract gait embeddings from IMU recordings in one internal dataset (ClinicalGait) and two external datasets (LTMM and FoG-STAR), and compare models fitted using covariates alone with those incorporating both covariates and gait embeddings.
Adding the embeddings increased AUROC on eight of nine tasks (Fig.~\ref{fig:application}b).
The largest gains were observed for FoG detection and ACL diagnosis, with AUROC increasing from approximately 0.25 to 0.98 and from 0.37 to 0.80, respectively.
These gains indicate that GaitFM extracts gait embeddings that complement demographic covariates, thereby providing reusable features for clinical assessment without task-specific fine-tuning.

\textbf{Retrieval of clinically similar patients.}
We used GaitFM to retrieve clinically similar cases from a reference database to support case-based clinical decision-making~\cite{wenckstern2026virtual} (Fig.~\ref{fig:application}c).
The pretrained GaitFM encoded each patient's IMU gait recordings into an embedding that represented their gait characteristics.
We built a searchable database by linking these embeddings to the corresponding clinical records in the reference cohort.
For a new patient, we computed the cosine distance between the query embedding and each reference embedding.
The system returned the cases with the smallest distances and their associated clinical records for clinician review.

To evaluate retrieval performance, we construct a PD case database using the mPower dataset and assess clinical similarity using five factors from the Movement Disorder Society-Unified Parkinson’s Disease Rating Scale (MDS-UPDRS)~\cite{goetz2008movement}, which capture symptoms and functional difficulties associated with PD.
Each factor is scored on a five-point ordinal scale ranging from 0 (normal) to 4 (severe).
For each factor, we calculate the average absolute score difference among the three nearest retrieved cases (MAE@3), averaged across queries.
GaitFM-based retrieval achieves a lower MAE@3 than random selection for all five factors (Fig.~\ref{fig:application}d), indicating greater clinical agreement between query patients and retrieved cases.
These results indicate that GaitFM embeddings capture gait characteristics associated with clinically relevant symptom profiles, supporting embedding-based retrieval of comparable cases for clinician review.
Detailed descriptions of the MDS-UPDRS table and retrieval results are provided in Supplementary Information Sections 3.2 and 4.8.

\subsection*{Deployment of GaitFM for on-device gait analysis}

\begin{figure*}[htbp]
	\centering
    \includegraphics[width=6.7in]{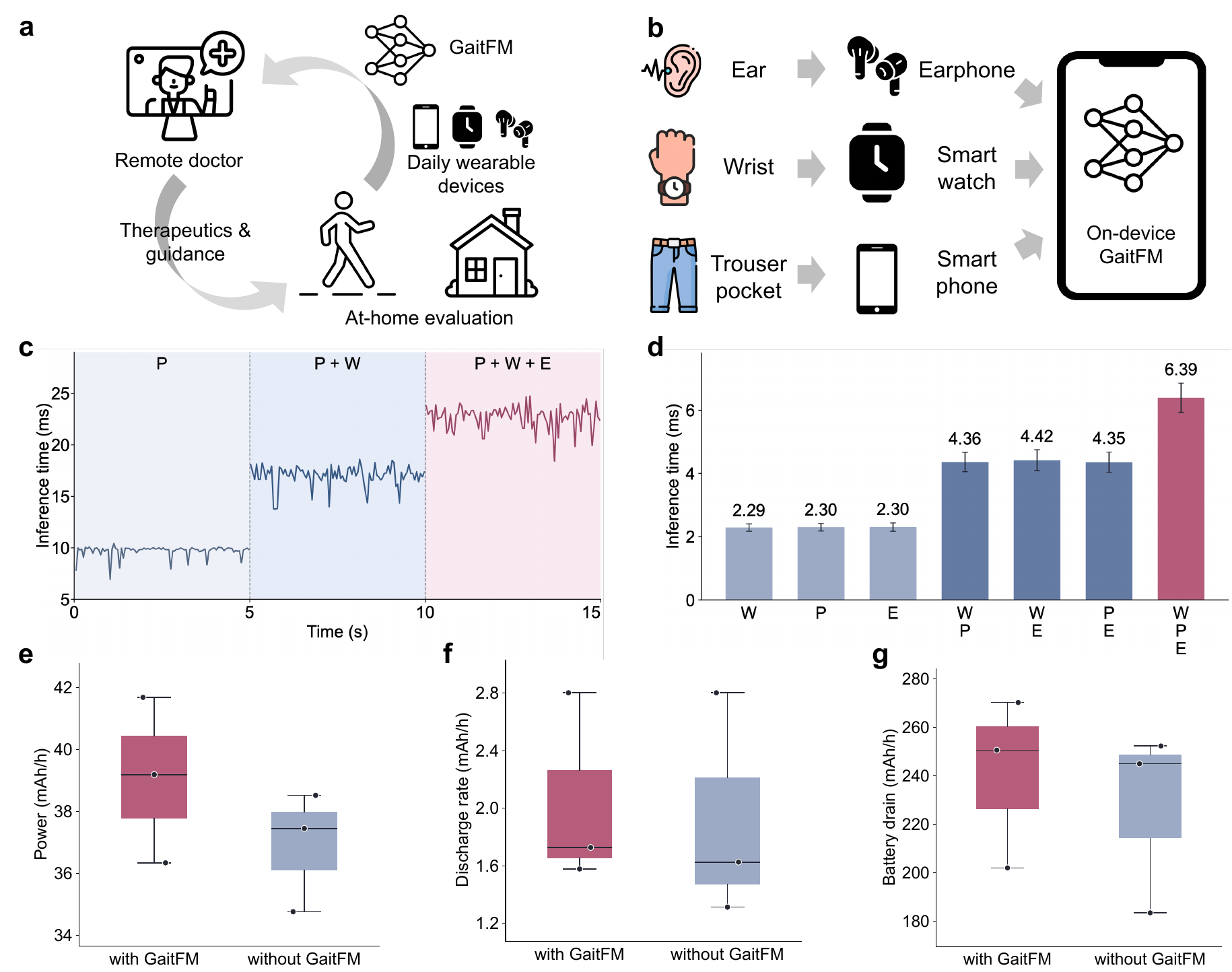}
	\caption{\textbf{GaitFM supports on-device gait assessment with millisecond-scale inference and modest additional battery drain.}
    \textbf{a,} Workflow for at-home gait assessment with remote clinical review and guidance.
    \textbf{b,} At-home gait assessment framework. 
    Everyday devices collect IMU data from the head, wrist, and thigh using earphone (E), a smartwatch (W), and a smartphone (P) carried in a trouser pocket, respectively. 
    Measurements from the earphone and smartwatch are transmitted wirelessly to the smartphone, where they are integrated with the smartphone measurements and processed locally by GaitFM.
    \textbf{c,} On-device inference latency during real-time IMU streaming with three configurations: smartphone alone (P,  $9.70\pm0.54$\ ms), smartphone and smartwatch (P + W, $16.98\pm1.02$ ms), and smartphone, smartwatch, and earphone (P + W + E, $22.64\pm1.12$ ms).
    \textbf{d,} Fixed-window inference latency across all seven device combinations. 
    Average inference latency ranges from 2.29 to 2.30\,ms for single-device inputs and from 4.35 to 4.42\,ms for two-device inputs, and reaches 6.39\,ms for three-device inputs.
    \textbf{e-g,} Additional energy consumption associated with GaitFM deployment for the smartwatch (\textbf{e}), earphones (\textbf{f}) and smartphone (\textbf{g}). 
    Each condition was evaluated in three independent runs.
    \textbf{e}, Deploying GaitFM increased the average battery discharge rate from 36.91 mAh\(\cdot\)h\(^{-1}\) to 39.07 mAh\(\cdot\)h\(^{-1}\) for the smartwatch, corresponding to an overhead of 5.85\%. 
    \textbf{f}, For the earphones, the average battery discharge rate increased from 1.914 mAh\(\cdot\)h\(^{-1}\) without GaitFM to 2.036 mAh\(\cdot\)h\(^{-1}\) with GaitFM, corresponding to an overhead of 6.39\%. 
    \textbf{g}, For the smartphone, the rate increased from 226.9 mAh\(\cdot\)h\(^{-1}\) to 240.9 mAh\(\cdot\)h\(^{-1}\), corresponding to an overhead of 6.2\%.}
	\label{fig:deploy}
\end{figure*}

IMUs embedded in everyday wearable devices, such as smartphones, smartwatches, and earphones, provide an opportunity for accessible, low-cost gait assessment outside specialized clinical facilities. 
Such a capability could be particularly valuable for individuals whose travel is restricted by gait impairment and for communities with limited access to specialist healthcare. 
Fig.~\ref{fig:deploy}a shows the proposed workflow for at-home gait assessment. 
Everyday wearable devices continuously collect gait data, which are analysed locally by GaitFM. 
The resulting gait assessments can be shared with remote clinicians, enabling clinical review and the delivery of therapeutic guidance without requiring patients to travel to a hospital or gait laboratory.

To evaluate the feasibility of this workflow, we deployed GaitFM on a smartphone (iPhone 17) and used IMU measurements from an earphone (AirPods 4), a smartwatch (Apple Watch Series 7), and the smartphone itself as model inputs (Fig.~\ref{fig:deploy}b). 
Measurements from the earphones and smartwatch are transmitted to the smartphone via Bluetooth, integrated with the smartphone measurements and processed locally by GaitFM.
These devices corresponded to sensors placed at the head, wrist, and trouser pocket, respectively. 
Under continuous IMU streaming, average inference latency was $9.70\pm0.54$ ms for the smartphone alone, $16.98\pm1.02$ ms for the smartphone and smartwatch, and $22.64\pm1.12$ ms for the smartphone, smartwatch and earphone (average $\pm$ s.d., Fig.~\ref{fig:deploy}c). 
We further measured fixed-window inference latency across all seven possible device combinations (Fig.~\ref{fig:deploy}d). 
Average latency ranged from 2.29 to 2.30\,ms for single-device inputs and from 4.35 to 4.42 ms for two-device inputs, reaching 6.39 ms for three-device inputs. 
These results demonstrate that GaitFM can process dynamic, multi-device IMU inputs on a consumer smartphone with millisecond-scale inference latency.
Deployment details are provided in Supplementary Information Section 3.3.

We also estimate battery discharge rates for the smartwatch, earphone, and smartphone during GaitFM deployment.
For the smartwatch, the average discharge rate increases from 36.91 to 39.07 mAh\(\cdot\)h\(^{-1}\), corresponding to a relative increase of 5.85\% (Fig.~\ref{fig:deploy}e).
For the earphone, the average discharge rate increases from 1.914 to 2.036 mAh\(\cdot\)h\(^{-1}\), corresponding to a relative increase of 6.39\% (Fig.~\ref{fig:deploy}f).
For the smartphone, the average discharge rate increases from 226.9 to 240.9 mAh\(\cdot\)h\(^{-1}\), corresponding to a relative increase of 6.2\% (Fig.~\ref{fig:deploy}g).
These results show that GaitFM adds little battery drain to daily wearable devices, supporting the feasibility of on-device gait analysis using daily wearable devices.

\section*{Discussion}

\subsection*{Implications}

Most existing IMU-based gait analysis models learn representations tailored to specific sensor configurations, body locations, or tasks, resulting in a fragmented landscape that limits their robustness and generalizability in diverse real-world settings. 
We introduce GaitFM, a unified foundation model specifically designed for IMU-based gait analysis.
GaitFM learns generalizable representations that transfer across tasks, sensor placements, and modalities, supporting gait assessment in settings ranging from clinical care to everyday life. 
Evaluated on 17 gait-related tasks (3 gait parameter estimation tasks, 13 disease and gait abnormality detection tasks, and 1 clinical outcome prediction task) across 11 heterogeneous datasets, GaitFM outperforms task- and location-specific end-to-end models.
GaitFM also supports flexible combinations of sensor inputs from different body locations and outperforms conventional models that rely on fixed sensor configurations. 
In summary, GaitFM offers a promising approach to addressing fragmentation in current digital health systems and therefore advancing the field.

Beyond improvements in predictive performance, the representations learned by GaitFM enable applications that extend the conventional prediction-centered paradigm of learning-based IMU gait analysis.
Conventional gait descriptors, such as gait speed, step length, and step time, have demonstrated clinical value across various conditions, but their reliable estimation from wearable IMUs can depend strongly on the population, recording environment, gait complexity, and analytical pipeline.
Pretrained GaitFM instead acts as a task-agnostic feature extractor, producing reusable representations that provide predictive information beyond demographic characteristics without task-specific adaptation.
These representations also capture gait characteristics associated with clinically relevant symptom profiles, enabling retrieval of patients with similar gait patterns from a reference database.
Together, these findings illustrate how foundation model representations can support applications beyond conventional prediction tasks and provide a basis for broader clinical applications of IMU-based gait analysis.

Clinically, a key practical advantage of GaitFM is its accessibility.
IMU data can be acquired using compact and relatively low-cost wearable sensors, in contrast to approaches that depend on cameras, force plates or specialized gait-laboratory infrastructure and therefore require assessment in instrumented environments.
Importantly, we evaluate the feasibility of (1) performing gait analysis using IMU data collected from body locations corresponding to everyday wearable devices (such as smartphone, smartwatch and earphone) and (2) using everyday wearable devices for IMU data acquisition and transmission, followed by on-device inference.
These results raise the possibility that future gait assessment could be performed using consumer devices already used in daily life, reducing dependence on customized IMU hardware.
If validated prospectively in real-world clinical populations, this approach could enable gait screening at population scale with minimal additional hardware cost.

Our experiments further suggest that general representation priors learned from the data-rich vision modality can be transferred to IMU data when an appropriate interface between the domains is established. 
In GaitFM, sensor-to-image representations and continued MAE pretraining enable visual knowledge to support gait assessment. 
Our results show that visual knowledge (1) provides powerful and transferable priors that help address sensor-data scarcity, (2) helps GaitFM capture local gait patterns and intersensor-modality relationships, and (3) improves the data efficiency of downstream gait-assessment tasks. 
We also show that (4) gait-assessment performance benefits from the scaling behavior of LVMs.
These findings point to a complementary strategy for alleviating sensor-data scarcity and improving generalization: identifying broadly reusable knowledge and developing effective ways to transfer it across modalities, beyond collecting large-scale datasets. 
This strategy may also provide a useful blueprint for other areas of sensor-based research, where physiological and behavioral sensing face similar constraints, such as limited labeled data and heterogeneous scenarios and devices.

\subsection*{Limitations}

This study has several limitations. 
First, the curated dataset does not cover the full spectrum of gait-related disorders. 
The included cohorts also differ in population characteristics, acquisition protocols, sensor hardware, and clinical definitions. 
Although this heterogeneity enables more comprehensive evaluation, it may introduce dataset-specific confounders and result in uneven representation across diseases and demographic groups. 
Future work should include additional neurological, musculoskeletal, and age-related disorders, with more balanced representation across age, sex, ethnicity, disease severity, and comorbidities. 
Second, the mechanisms underlying cross-modal transfer remain incompletely understood. 
It is not yet clear which visual priors are transferred to sensor representations or why they are beneficial. 
Future studies can improve the interpretability of the transfer process and identify the visual knowledge that contributes to sensor-data understanding. 
Despite this uncertainty, our complementary analyses provide initial evidence from multiple perspectives that visual knowledge can improve the understanding of gait-related sensor data, laying a foundation for further research. 
Third, this study provides an early proof of concept and establishes the technical feasibility of GaitFM. 
Clinical translation will require prospective studies to evaluate its performance and utility in real-world settings.

Despite these limitations, this study represents a notable advance in IMU-based human gait analysis by improving performance across gait-related tasks, enabling flexible use of sensor configurations, and reducing reliance on large-scale sensor datasets.
The learned embeddings further support applications including zero-shot feature extraction and retrieval of clinically similar cases.
Together, these contributions may advance the development of accessible, reliable and low-cost remote gait assessment.

\section*{Methods}

\subsection*{Dataset and preprocessing}

We curated 11 IMU-based gait datasets comprising 4,117 participants. 
The datasets were heterogeneous in sensor modality (accelerometer only or accelerometer and gyroscope), recorded gait activities (differing in walking types), device type (research-grade IMUs and IMUs embedded in consumer electronics), sensor placement combinations (1-14 locations), populations (differing in age, sex, and health status), and acquisition protocol (free-living, predefined tasks, and controlled lab settings). 

To harmonize data across datasets, all signals were resampled to 50 Hz, divided into non-overlapping 5-s windows and standardized using z-score normalization. 
This preprocessing yielded approximately 1.41 million single-location windows.
For datasets containing simultaneous recordings from multiple locations, we enumerated all available combinations of sensor locations. 
For datasets providing both accelerometer and gyroscope measurements, we randomly masked the gyroscope channels with a probability of 0.5 during pretraining and training, exposing the model to both accelerometer-only and multimodal inputs.
These procedures exposed the model to varying numbers and placements of sensors, allowing it to accommodate different sensor configurations without retraining.

Participants were assigned to training, validation and test sets in a 6:2:2 ratio, with all samples from each participant restricted to a single partition to prevent data leakage. 
The training partitions of the seven internal datasets were used for both model pretraining and downstream fine-tuning. 
For the four external datasets, the training partitions were used only for downstream fine-tuning.

\subsection*{GaitFM architecture and pretraining}

\textbf{Sensor-to-Image embedder.}
We first transform each raw IMU window into an image-like representation using a sensor-to-image embedder. Let $\mathbf{X}\in\mathbb{R}^{C\times L}$ denote an IMU window, where $C$ and $L$ represent the number of sensor channels and sequence length, respectively. Each sensor channel is treated as an independent univariate sequence by reshaping the input from $C\times L$ to $C\times1\times L$.

A shared one-dimensional convolution is then applied independently to each sensor channel. The convolution uses $s=10$ for both its kernel size and stride and produces $3s$ output channels. Its weights are constrained to be non-negative and to sum to one along the temporal dimension. Specifically,
\begin{equation}
z_{c,q,n}
=
\sum_{j=0}^{s-1}
\bar{w}_{q,j}\,
\tilde{x}_{c,ns+j},
\qquad
\bar{w}_{q,j}
=
\frac{|w_{q,j}|}
{\sum_{j'=0}^{s-1}|w_{q,j'}|+\epsilon},
\label{eq:sensor_temporal_embedding}
\end{equation}
where $\tilde{x}$ denotes the same-padded input, $q\in\{0,\ldots,3s-1\}$ indexes the convolutional output channels, and $n\in\{0,\ldots,N-1\}$ indexes the temporal segments, with $N=\lceil L/s\rceil$. The resulting temporal embedding is denoted by $\mathbf{Z}\in\mathbb{R}^{C\times(3s)\times N}$. Because the convolutional filters are shared across all sensor channels, the same temporal transformation is applied to each channel.
We then rearrange $\mathbf{Z}$ into a three-channel image tensor $\mathbf{I}\in\mathbb{R}^{3\times(Cs)\times N}$:
\begin{equation}
I_{r,cs+u,n}
=
z_{c,rs+u,n},
\qquad
r\in\{0,1,2\},\quad
u\in\{0,\ldots,s-1\}.
\label{eq:sensor_to_image}
\end{equation}
The resulting representation is bilinearly resized to $3\times224\times224$, independently min-max normalized within each image channel and mapped to the range $[-1,1]$.

\textbf{Sensor-masked pretraining.}
We continue pretraining a Vision Transformer (ViT)~\cite{he2022masked} initialized with ImageNet-1K-pretrained weights using the sensor-derived images. 
Each sensor image is divided into non-overlapping $16\times16$ patches, of which 75\% are randomly masked, leaving only the visible tokens to be processed by the encoder. 
This high masking ratio limits reliance on locally redundant information and encourages the model to reconstruct the missing content by learning complex dependencies among the visible tokens. 
Moreover, processing only 25\% of the tokens substantially reduces the computational cost of pretraining.

\textbf{Sensor encoder.}
The patch-embedding layer flattens each image patch $p_i$ and maps it through a shared linear projection into a $D$-dimensional token embedding. 
Positional embeddings are added to each token to preserve its spatial position within the sensor image. 
Following masking, a learnable classification token $\mathbf{e}_{\mathrm{CLS}}$ is prepended to the visible token sequence, which is then processed by a stack of Transformer encoder blocks. 
Each block comprises multi-head self-attention and a feed-forward network with residual connections and layer normalization, allowing the encoder to capture dependencies across sensor channels and temporal regions. 
The encoder produces a classification token $\mathbf{z}_{\mathrm{CLS}}$ and a set of sensor tokens $\mathbf{Z}_{\mathrm{sensor}}=\{\mathbf{z}_i\}_{i=1}^{M}$.

\textbf{Sensor decoder.}
The sensor decoder reconstructs the masked sensor-image patches from the encoded visible tokens. 
It first maps the encoder output into the decoder embedding space through a linear projection. 
A learnable mask token is then inserted at each masked position, and the complete token sequence is restored to its original spatial order. 
Positional embeddings are added to the reordered tokens, which are subsequently processed by a stack of Transformer decoder blocks followed by layer normalization. 
A final linear projection maps each decoded sensor token to a predicted patch $\hat{\mathbf{p}}_i\in\mathbb{R}^{d_p}$, where $d_p$ is the number of pixel values within a patch. The classification token is excluded from patch prediction.

For each target patch $\mathbf{p}_i$, we normalize its elements using the patch-specific mean $\mu_i$ and variance $\sigma_i^2$. 
The reconstruction loss is computed only over the set of masked patches $\mathcal{M}$:
\begin{equation}
\tilde{\mathbf{p}}_i
=
\operatorname{sg}\!\left(
\frac{\mathbf{p}_i-\mu_i}
{\sqrt{\sigma_i^2+\epsilon}}
\right),
\qquad
\mathcal{L}_{\mathrm{rec}}
=
\frac{1}{|\mathcal{M}|}
\sum_{i\in\mathcal{M}}
\frac{1}{d_p}
\left\|
\hat{\mathbf{p}}_i-\tilde{\mathbf{p}}_i
\right\|_2^2,
\label{eq:sensor_reconstruction}
\end{equation}
where $\operatorname{sg}(\cdot)$ denotes the stop-gradient operation and $\epsilon$ is a small constant for numerical stability. The sensor decoder is used only during pretraining.

\subsection*{Adaptation to downstream tasks}

Two key capabilities of foundation models are 
(i) learning generalizable representations from raw data that can be transferred across downstream tasks and 
(ii) accommodating flexible input formats. 
To evaluate these capabilities, we assess the pretrained GaitFM across multiple clinical tasks and dynamic sensor inputs.

\textbf{Dynamic-input fine-tuning.}
For downstream evaluation, we discard the sensor decoder and retain the pretrained sensor-to-image embedder and sensor encoder, both of which are updated through full-parameter fine-tuning. 
Each available sensor location is processed independently by the shared sensor-to-image embedder and sensor encoder. 
A task-specific location representation $\mathbf{h}_k$ is extracted from the encoder outputs and combined with a learnable location embedding $\mathbf{e}_k$ to preserve anatomical information. 
The resulting representations are assigned to fixed slots according to their sensor locations, while slots corresponding to unavailable locations are filled with zeros. 
A binary availability mask $\mathbf{m}$ indicates which slots contain valid inputs. 
The fixed-slot representations are flattened and concatenated with $\mathbf{m}$ to form a fixed-dimensional representation $\mathbf{h}_{\mathrm{dyn}}$, which is passed to the corresponding task-specific prediction head. 
This design accommodates different numbers and combinations of sensor locations within a unified architecture.

\textbf{Task-specific adaptation of pretrained models.}
We consider downstream tasks spanning three clinical domains: gait parameter estimation, disease and abnormality diagnosis, and clinical outcome prediction. 
Gait parameter estimation is formulated as a regression task, whereas disease and abnormality diagnosis and clinical outcome prediction are formulated as classification tasks.
Detailed task descriptions are provided in Supplementary Information Section 3.1.
We use distinct task-specific adaptation strategies for these two task categories, as described below.

\textit{Classification tasks.}
For each available sensor location, the CLS token $\mathbf{z}_{\mathrm{CLS},k}$ serves as the sensor representation $\mathbf{h}_k$.
A learnable location embedding is added to each representation, after which the location-aware representations are assigned to fixed slots and concatenated with a binary availability mask to form the dynamic-input representation $\mathbf{h}_{\mathrm{dyn}}$.
An MLP classifier maps $\mathbf{h}_{\mathrm{dyn}}$ to $K$ class logits.
The sensor-to-image embedder, sensor encoder, location embeddings and MLP classifier are jointly optimized using the cross-entropy loss.

\textit{Regression tasks.}
For regression tasks, the sensor tokens $\mathbf{Z}_{\mathrm{sensor},k}$ from each available sensor location are processed by a shared ResNet to obtain the corresponding location representation $\mathbf{h}_k$.\
For multi-location inputs, a learnable location embedding is added to each representation, and the resulting representations are assigned to fixed slots and concatenated with a binary availability mask to form the dynamic-input representation $\mathbf{h}_{\mathrm{dyn}}$.\
The resulting representation is mapped to a continuous prediction $\hat{y}$ by the regression head.\
The sensor-to-image embedder, sensor encoder, ResNet, location embeddings and regression head are jointly optimized using the smooth $L_1$ loss.

\subsection*{Experimental settings}

\textbf{Implementation of pretraining.}
GaitFM is pretrained for 5 epochs on NVIDIA RTX PRO 6000 GPUs using the Adam optimizer with a global batch size of 2048. 
The initial learning rate is set to $1\times10^{-5}$ and decayed by a factor of 0.5 according to the type-1 learning-rate schedule. 
We use $\beta_1=0.9$, $\beta_2=0.999$ and no weight decay.

\textbf{Implementation of downstream tasks.}
For regression tasks, the end-to-end baseline models use a learning rate of $1\times10^{-4}$ and a batch size of 64. 
GaitFM uses separate learning rates of $1\times10^{-5}$ for the sensor-to-image embedder and sensor encoder and $1\times10^{-4}$ for the regression head, with a batch size of 128. 
For classification tasks, the end-to-end baseline models use a learning rate of $1\times10^{-4}$ and a batch size of 128. 
GaitFM uses learning rates of $1\times10^{-5}$ for the sensor-to-image embedder and sensor encoder and $1\times10^{-4}$ for the MLP classification head and location embeddings, with a batch size of 128. 
All models are optimized with AdamW and trained on NVIDIA RTX PRO 6000 GPUs.

\subsection*{Evaluation metrics}

For classification tasks, we evaluate performance using the area under the receiver operating characteristic curve (AUROC) at the subject level. 
For FoGContest, we use sample-level AUROC to reflect symptom-event monitoring. 
We also use sample-level AUROC for FoG-STAR and StrokeRehabilitation because these datasets include few subjects. 
For regression tasks, we report sample-level mean absolute error (MAE) and the coefficient of determination (\(R^2\)). 
Error bars denote 95\% CIs.

\subsection*{Statistical analysis}

The 95\% CIs were estimated using non-parametric bootstrapping with 1,000 resamples. 
For regression tasks, pairwise differences between models were assessed using paired bootstrap tests with 1,000 resamples. 
For classification tasks, pairwise differences were assessed using two-sided Wilcoxon signed-rank tests. 
Values of $P<0.05$ were considered statistically significant, with $P<0.01$ and $P<0.001$ reported where applicable.

\subsection*{Data availability}

The ImageNet-1K dataset~\cite{deng2009imagenet} used for vision pretraining is available online (\url{https://huggingface.co/datasets/ILSVRC/imagenet-1k}). 
The sensor pretraining and internal validation datasets, GSTRIDE~\cite{garcia2022gstride} (\url{https://zenodo.org/records/17052815}), FoGContest~\cite{salomon2024machine} (\url{https://zenodo.org/records/10959560}), COMPWALK-ACL~\cite{yona2025compwalk} (\url{https://zenodo.org/records/15624356}), mPower~\cite{bot2016mpower} (\url{https://www.synapse.org/Synapse:syn4993293/wiki/}), WearGait-PD~\cite{anderson2024weargait} (\url{https://www.synapse.org/Synapse:syn52540892/wiki/623751}), Mobilise-D~\cite{mazza2021technical} (\url{https://zenodo.org/records/13899386}), and ClinicalGait~\cite{voisard2025dataset} (\url{https://doi.org/10.6084/m9.figshare.28806086}) are all available online. 
The external validation datasets LTMM~\cite{weiss2013does} (\url{https://physionet.org/content/ltmm/1.0.0/}), FoG-STAR~\cite{borzi2026multi} (\url{https://zenodo.org/records/17838806}), StrokeRehabilitation~\cite{zhou_2024_10534055} (\url{https://zenodo.org/records/10534055}), and GaitPrint~\cite{wiles2025nonan} (\url{https://springernature.figshare.com/articles/dataset/NONAN_GaitPrint_An_IMU_gait_database_of_healthy_middle-aged_adults/29371796}) are all available online as well.

\subsection*{Code availability}
The source code is available on GitHub at \url{https://github.com/saiketa/GaitFM}.
An online demo to test GaitFM is available at \url{https://huggingface.co/spaces/OnKIKU/gaitfm-fogcontest-demo}.

\bibliographystyle{unsrtnat}
\bibliography{references}

\end{document}